\documentclass[runningheads]{llncs}

\usepackage{eccv}
\RequirePackage[width=122mm,left=12mm,paperwidth=146mm,height=193mm,top=12mm,paperheight=217mm]{geometry}

\usepackage{eccvabbrv}

\usepackage{graphicx}
\usepackage{booktabs}

\usepackage[accsupp]{axessibility}  

\usepackage{hyperref}

\usepackage{orcidlink}

\usepackage[linesnumbered,ruled,vlined]{algorithm2e}
\usepackage{rotating}
\usepackage{pifont}
\usepackage[mathscr]{eucal}
\usepackage{array,multirow}
\usepackage{tabularx}
\usepackage{bbm}
\newcolumntype{b}{X}
\newcolumntype{s}{>{\hsize=.5\hsize}X}
\definecolor{fgreen}{rgb}{0.13, 0.55, 0.13}
\definecolor{bred}{rgb}{0.55, 0.13, 0.13}
\definecolor{plot1}{HTML}{1f77b4}

\definecolor{plot2}{HTML}{ff7f0e}

\definecolor{plot3}{HTML}{2ca02c}

\DeclareMathOperator*{\argmax}{argmax}

\usepackage{wrapfig}
\newcommand{\vect}[1]{\mathbf{#1}}

\begin{document}

\title{Four Ways to Improve Verbo-visual Fusion for Dense 3D Visual Grounding} 

\titlerunning{Four Ways to Improve Verbo-visual Fusion for Dense 3D Visual Grounding}

\author{Ozan~Unal\inst{1,2}\orcidlink{0000-0002-1121-3883} \and
Christos~Sakaridis\inst{1} \and
Suman~Saha\inst{1,3} \and Luc~Van~Gool\inst{1,4,5}}

\authorrunning{O.~Unal et al.}

\institute{$^1$ETH Zurich, $^2$Huawei Technologies, $^3$PSI, $^4$KU Leuven, $^5$INSAIT \\
\email{\{ozan.unal, csakarid, suman.saha, vangool\}@vision.ee.ethz.ch}}
\maketitle

\begin{abstract}
\vspace{-10px} 3D visual grounding is the task of localizing the object in a 3D scene which is referred by a description in natural language. With a wide range of applications ranging from autonomous indoor robotics to AR/VR, the task has recently risen in popularity. A common formulation to tackle 3D visual grounding is grounding-by-detection, where localization is done via bounding boxes. However, for real-life applications that require physical interactions, a bounding box insufficiently describes the geometry of an object. We therefore tackle the problem of \textit{dense} 3D visual grounding, i.e. referral-based 3D instance segmentation. We propose a dense 3D grounding network ConcreteNet, featuring four novel stand-alone modules that aim to improve grounding performance for challenging repetitive instances, i.e. instances with distractors of the same semantic class. First, we introduce a bottom-up attentive fusion module that aims to disambiguate inter-instance relational cues, next, we construct a contrastive training scheme to induce separation in the latent space, we then resolve view-dependent utterances via a learned global camera token, {and finally we employ multi-view ensembling to improve referred mask quality}. ConcreteNet ranks $1^{st}$ on the challenging ScanRefer online benchmark and has won the ICCV $3^{rd}$ Workshop on Language for 3D Scenes ``3D Object Localization'' challenge. Our code is available at \url{ouenal.github.io/concretenet/}.
\end{abstract}

\setlength\intextsep{0pt}
\begin{wrapfigure}{r}{0.49\textwidth}
    \centering
    \vspace{-25px}
    \includegraphics[width=.48\textwidth]{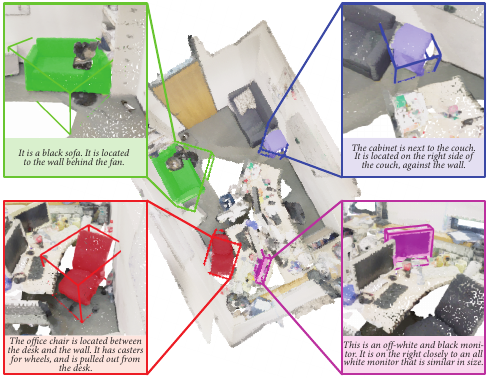}
     \caption{{ConcreteNet localizes referred objects via dense masks rather than boxes.}}
     \label{fig:teaser} 
\end{wrapfigure}

\section{Introduction}
\label{sec:intro}

As the field of natural language processing (NLP) continues to mature and develop, the quest to mimic language-driven human-to-human interactions in how AI interacts with humans is beginning. In the case where AI agents are embodied and need to interact with humans inside a real 3D environment with rich visual information, \emph{grounding} language to this environment becomes of utmost importance for understanding human utterances, which is, in turn, the sine qua non for the successful operation of such agents.

3D visual grounding is the task of localizing an object within 3D space based on a natural language prompt, i.e., referral-based 3D object localization. Compared to its 2D counterpart which has been widely studied~\cite{kazemzadeh2014referitgame, mao2016generation, plummer2015flickr30k}, designing robust models exploiting language - 3D point cloud multi-modality remains a highly complex and challenging task.
Attention to 3D visual grounding has recently risen thanks to the introduction of the cornerstone datasets ScanRefer~\cite{chen2020scanrefer}, Nr3D and Sr3D~\cite{achlioptas2020referit3d}. These datasets are based on the 3D ScanNet~\cite{dai2017scannet} dataset and additionally contain either free-form~\cite{chen2020scanrefer} or contrastive~\cite{achlioptas2020referit3d} lingual descriptions, each of which refers to a single 3D object in the scene which must be recognized. This subtle difference in the construction of descriptions between ScanRefer on the one side and Nr3D and Sr3D on the other induces two respective variants of 3D visual grounding. The former variant consists of the 3D localization of the referred object directly from the point cloud~\cite{chen2020scanrefer}, while the latter additionally provides as input the ground-truth 3D bounding boxes of all objects in the scene which belong to the class of the referred object~\cite{achlioptas2020referit3d}. We refer to the former variant as referral-based 3D object localization and to the latter as referral-based 3D object identification. Referral-based 3D object localization is arguably more challenging, as it requires (i) \emph{detecting} multiple candidate objects from several classes, including classes \emph{besides} that of the referred object, and (ii) discriminating between the referred object and \emph{all} other candidate objects. As our goal is to achieve an end-to-end solution to verbo-visual fusion, we focus on this variant.

State-of-the-art methods~\cite{chen2020scanrefer,zhao20213dvgt, chen2022ham} focus on grounding descriptions to 3D \emph{bounding boxes} of referred objects.
While such detection-level grounding has vast potential for real-world applications, ranging from autonomous robots to AR/VR, the level of geometric detail which is provided by 3D bounding box detections remains limited. As an example, autonomous robots may have to grasp or avoid objects. Therefore, delivering a detailed, pointwise mask is more beneficial for downstream tasks than just having the axis-aligned bounding boxes. We illustrate this in Fig.~\ref{fig:teaser}.

Considering the above-mentioned goals, this paper tackles the underexplored problem of \emph{dense} 3D visual grounding~\cite{huang2021tgnn,yuan2021instancerefer}. We propose a novel dense 3D visual grounding network (ConcreteNet), where we adopt the commonly used grounding-by-selection strategy. In grounding-by-selection, a visual backbone first produces 3D instance candidates with a point cloud as input. After that, a verbo-visual fusion module selects the correct candidate based on the natural language description. With this framework, we observe that 3D instance segmentation yields more robust and tighter predictions compared to 3D object detection, but suffers from reduced separability in the latent space between instances of the same semantic class. This results in a significant increase in false positive rates for the higher level 3D visual grounding task, most notably observed for repetitive instances, i.e. instances that are not semantically unique in a scene. To combat this effect, we propose four novel ways to improve verbo-visual fusion for dense 3D visual grounding.

First, we observe that in cases where referrals may be construed as valid for multiple instances, locality rules. In other words, due to our limited attention spans, we humans mainly consider nearby objects when referring to an instance. To disambiguate inter-instance relational cues, we propose a bottom-up attentive fusion module (BAF) that induces this locality during verbo-visual fusion via spherical masking with an increasing radius, allowing only neighboring objects to attend to each other.

Second, we form a general solution to the instance separability issue within the latent verbo-visual space by constructing a contrastive training scheme to alleviate ambiguities between embeddings of repetitive instances. Specifically, we pull sentence embeddings and matching instance vectors towards each other, while contrasting non-matching pairs.

Next, we tackle the issue of view-dependent referrals. Unlike in 2D, 3D scenes do not inherently possess a directional right or left side, or a room does not have a clear back or front. However, often our perception is unequivocally guided by our personal perspectives, and thus such view-dependent descriptions are unavoidable in any real-world situation. While we can empathize with the speaker and rationalize the possible viewpoint associated with an utterance, this trait does not come naturally to machines. We therefore propose to introduce a learned global camera token (GCT) that can be directly supervised via the camera positions used during annotation to help resolve view-dependent prompts.

{Finally, we improve the quality of our predicted referred instance mask through ensembling over multiple views of a single point cloud scene by reducing the epistemic uncertainty. To this end, we develop a two-stage ensemble algorithm for dense 3D visual grounding that first determines the correct referred object from all predictions of the individual viewpoints and then refines the aggregated instance mask.}

\noindent In summary, our contributions are as follows:
\begin{itemize}
    \item We present ConcreteNet, a kernel-based 3D visual grounding network that predicts 3D instance masks as opposed to 3D bounding boxes to aid real-world applications that require a fine geometric understanding of an object.
    \item We introduce a bottom-up attentive fusion module (BAF) to disambiguate inter-instance relational referrals through spherical neighborhood masking.
    \item We construct a contrastive learning scheme to induce further separation in the latent representation to aid repetitive instance grounding.
    \item We learn a global camera token (GCT) to resolve view-dependent prompts.
    \item We propose multi-view ensembling to improve referred mask quality.
\end{itemize}
With all four proposed improvements, we rank $1^{st}$ in the challenging ScanRefer~\cite{chen2020scanrefer} online benchmark.

\section{Related Work}
\label{sec:related}

\noindent\textbf{3D visual grounding} is a prominent 3D task in the area of vision and language and constitutes the 3D version of the more extensively studied task of 2D visual grounding~\cite{kazemzadeh2014referitgame,mao2016generation,plummer2015flickr30k}, which aims to ground a verbal description for an image to the specific object this description refers to. Respectively, 3D visual grounding methods~\cite{prabhudesai2020embodied,feng2021free,luo20223d} accept a 3D scene in the form of a point cloud as visual input and need to ground the accompanying description to the referred object in 3D. Closely related 3D vision-language tasks are 3D dense captioning~\cite{chen2021scan2cap,chen2022d3net} and grounding spatial relations (instead of individual objects) in 3D~\cite{goyal2020rel3d}.

\sloppy{An early, seminal work in 3D visual grounding~\cite{kong2014coreference} employed an MRF to densely ground lingual descriptions of 3D scenes from the NYU Depth-v2 dataset~\cite{NYU}, which reasons jointly about the 3D scene and its description.
Attention modules were proposed in~\cite{zhao20213dvgt} both for leveraging context in the object proposal module and for the verbo-visual fusion module. We build our attentive verbo-visual fusion module on top of the one used in~\cite{zhao20213dvgt}, but we propose four novel ways to improve this fusion. In particular, we improve (i) the internal fusion mechanism by enforcing progressive context aggregation across object candidates via masked attention, (ii) the supervision of the fusion outputs via a cross-modal contrastive loss which uniquely attracts the visual embedding of the referred object to its verbal embedding, (iii) the sensitivity of object embeddings to the viewpoint from which the verbal description is generated by including a dedicated, learned global camera token in our attentive fusion and finally (iv) the referred mask predictions by lowering the epistemic uncertainty through multi-view ensembling}.


Previous 3D grounding works have explored local attention~\cite{chen2022ham} and viewpoint dependency~\cite{huang2022multiview,roh2022languagerefer}, similarly to us. 3DVG-T~\cite{zhao20213dvgt} utilizes relative Euclidean distance for relational encoding. While this helps capture object-to-object interactions, the model still relies on global attention for information routing. Extending 3DVG-T, 3DJCG~\cite{cai20223djcg} introduces an additional spatial distance matrix, computed from the centers of the initial object proposals. While this matrix acts as a relation encoder on the attention maps, the model still relies on a global attention scheme where all object tokens can exchange information. By contrast, we induce hard locality through spherically masked attention in a bottom-up manner. Rendering attention local, i.e., only allowing object-to-object message passing between neighbors, helps the model better disambiguate inter-object relations and improve the 3DG performance for the \emph{multiple} cases.
Chen~\etal.~\cite{chen2022ham} provide language embeddings of multiple granularities as inputs to verbo-visual fusion, which includes a module implementing local attention via partitioning the 3D volume of the scene into coarse voxels and restricting attention across different visual embeddings within each voxel. By contrast, our attentive fusion implements locality in an isotropic fashion, using spherical attention masks centered at the centroid of the respective object. MVT~\cite{huang2022multiview} addresses the insensitivity of vision-based object embeddings to the description viewpoint in a data-driven fashion by applying rotation augmentations to the input 3D scene. Instead of proliferating the already sizable 3D inputs of the grounding model, we inject viewpoint sensitivity in our verbo-visual attentive fusion by including an additional, camera viewpoint token in the visual tokens of our attention, which induces a comparatively negligible computational overhead. The empirical findings of~\cite{roh2022languagerefer} support the positive effect of even approximate viewpoint information on referral-based 3D object identification accuracy on Nr3D scenes with view-dependent descriptions. This information is provided in~\cite{roh2022languagerefer} by canonicalizing the yaw of the scenes with respect to their descriptions. We instead \emph{learn} the viewpoints of ScanRefer data from exact annotations as part of our verbo-visual fusion, leveraging the abundant viewpoint-dependent descriptions in these data.

\noindent\textbf{Dense 3D visual grounding} or referral-based 3D instance segmentation presents the additional challenge of precisely segmenting the 3D points belonging to the referred object from points belonging to other objects or to the background, and it is far less explored than standard 3D detection-level grounding. In Huang~\etal~\cite{huang2021tgnn}, verbo-visual fusion between instance embeddings and word embeddings is performed with a graph neural network. The attention-based verbo-visual fusion of~\cite{yuan2021instancerefer} only accepts the global sentence embedding as input, which does not allow the instance embeddings to attend to individual words that may carry more specific information about geometry and appearance. Semantic instance-specific features produced in~\cite{yuan2021instancerefer} in the process of extracting the candidate instances are discarded in the subsequent extraction of instance embeddings for grounding, whereas we learn these semantic features end-to-end, optimizing them both for the generation of instance embeddings that are discriminative for grounding and for segmentation accuracy.

\noindent\textbf{Verbo-visual contrast} has been shown to provide a strong alternative to traditional categorical visual supervision for learning discriminative 2D visual representations. In particular, CLIP~\cite{radford2021clip} learns a multi-modal, verbo-visual embedding space by contrasting 2D visual embeddings to language embeddings based on the co-occurrence of respective inputs from the two modalities. Our proposed contrastive loss for 3D visual grounding also applies verbo-visual contrast between the embedding of the verbal description and the visual 3D \emph{object-level} embeddings, which effectively pushes the embedding of the referred object away from the embeddings of other objects and thus aids classification. Another work that leverages verbo-visual contrast in a 3D task is~\cite{rozenberszki2022language}, which contrasts learned 3D semantic segmentation features with features from a pre-trained CLIP model based on the class of the respective 3D and verbal inputs.

\noindent{\textbf{Multi-view in 3D visual grounding} is leveraged in MVT~\cite{huang2022multiview} which aggregates the features from multiple views to reduce dependence on specific views, and in ViewRefer~\cite{guo2023viewrefer} that also utilizes multi-view text input. Multi3DRefer~\cite{zhang2023multi3drefer} generates multi-view 2D images from 3D objects and employs a CLIP image encoder to inject rich features into object candidates. Compared to the aforementioned works, our multi-view ensembling (MVE) directly operates on selected referred objects rather than each individual predicted object, i.e. acts as late multi-view aggregation rather than early multi-view fusion. Thus not only does MVE reduce the epistemic uncertainty within the selection (similar to previous works) but also within the final mask prediction. In other words, our method does not fully rely on an initial object proposal to determine the final mask, but further uses the multi-view information to construct a refined mask.
}

\begin{figure}[t]
\centering
    \includegraphics[width=\textwidth]{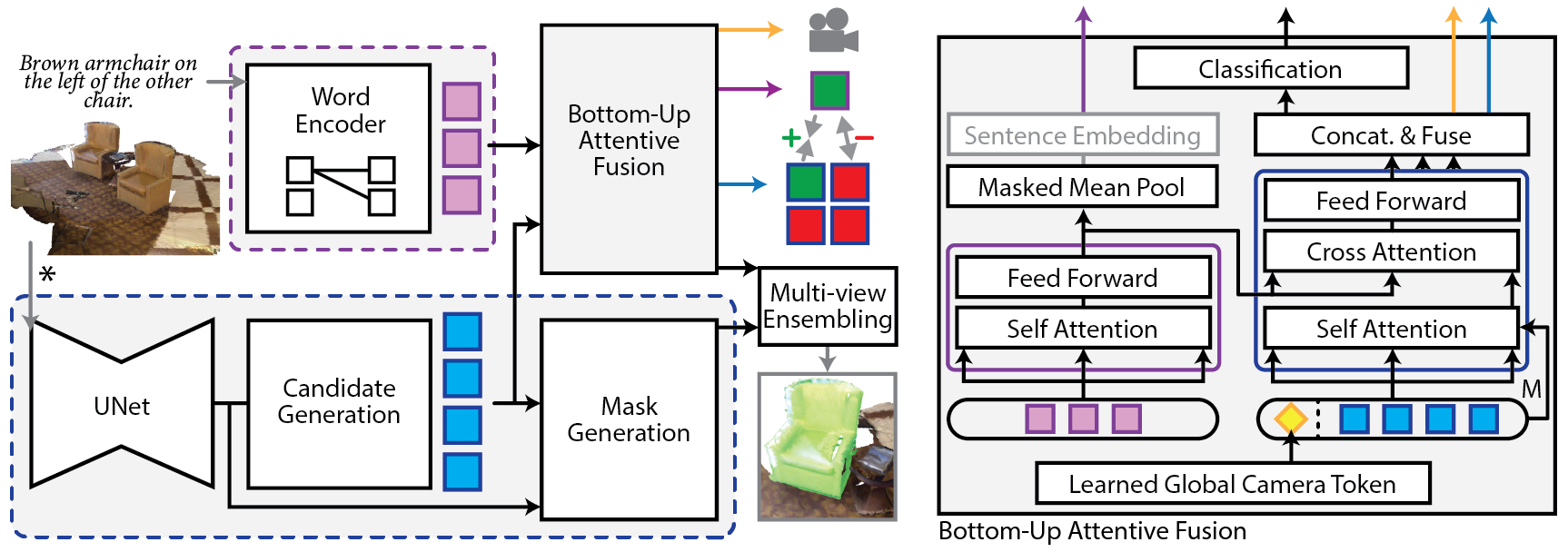}
    \caption{Illustration of our ConcreteNet dense 3D visual grounding pipeline (left). Given a point cloud and a natural language prompt, we first generate instance candidates (blue) and word embeddings (pink). We then fuse these to densely ground the verbal description to the 3D scene. We improve performance by localizing attention via a bottom-up attentive fusion module (right), utilizing contrastive learning to promote better feature separability, and learning the camera position to disambiguate view-dependent descriptions. Our final prediction is generated by merging the token of the best-fitting instance with its predicted mask.}
    \label{fig:pipeline}
\vspace{-10px} \end{figure}

\vspace{-10px}\section{Method}\vspace{-10px}


Our architecture comprises three core modules: a visual encoder,
a verbal encoder and a verbo-visual fusion module. In Sec.~\ref{sec:instance}, we introduce the 3D visual backbone which generates 3D instance candidates along with the ensuing masks. In Sec.~\ref{sec:language}, we outline how we encode the language queries into high-dimensional word embeddings, and in Sec.~\ref{sec:grounding}, we present our verbo-visual fusion module for grounding a description in 3D space by fusing the word embeddings and instance embeddings to predict the referred instance mask. The overall pipeline is seen in Fig.~\ref{fig:pipeline} - left.

\subsection{Kernel-Based 3D Instance Segmentation}
\label{sec:instance}

3D point clouds are large, unordered data structures. Commonly, dense tasks such as instance segmentation require a dense representation, thus high-level feature information needs to be captured in relatively high resolution~\cite{jiang2020pointgroup, vu2022softgroup}. Following attentive verbo-visual fusion approaches~\cite{zhao20213dvgt, chen2022ham},  interactions between language cues and a sizeable number of points result in a considerable amount of computing and memory requirements. By contrast, kernel-based instance segmentation models condense instance information within a single scene into a sparse representation in the form of instance-aware kernels~\cite{wu2022dk, he2021dyco3d}. These kernels are then used to scan the whole scene to reconstruct instance masks via dot product or dynamic convolution. Our kernel-based 3D instance segmentation pipeline is illustrated in Fig.~\ref{fig:pipeline} - blue.

Formally, following recent literature~\cite{vu2022softgroup}, we first extract features $\vect{f}_{\text{3D}} \in \mathbb{R}^{N\times{}D}$ for all $N$ 3D points using a sparse convolutional UNet backbone~\cite{jiang2020pointgroup}. The resulting features are then used to predict an auxiliary semantic prediction $\vect{s} \in \mathbb{R}^{N \times C}$, where $C$ is the number of classes, and the offsets $\vect{x} = \vect{p} - \vect{o} \in \mathbb{R}^{N\times{}3}$ from points $\vect{p}$ to their instance centroids $\vect{o}$. We supervise via:
\begin{equation}
    \mathcal{L}_{\text{sem}} = H(\vect{s}, \hat{\vect{s}}) \textrm{\phantom{sp} and \phantom{sp}} \mathcal{L}_{\text{off}} = L_1(\vect{x}, \hat{\vect{x}}),
\end{equation}
with $H$ and $L_1$ denoting the cross-entropy and $L_1$ losses respectively, and \, $\hat \cdot$ \, denoting ground-truth values. For the offset loss $\mathcal{L}_{\text{off}}$, we ignore points that do not belong to an associated instance.

\noindent\textbf{Candidate generation.} To generate instance candidates from pointwise features, we closely follow DKNet~\cite{wu2022dk}. We generate a sharp centroid map $h$ by concatenating $\vect{f}_{\text{3D}}$ and $\vect{o}$ and processing the joint information via an MLP and an ensuing softmax operation. The centroid maps are supervised via geometry-adaptive Gaussian kernels applied to ground-truth heatmaps:
\begin{equation}
    \mathcal{L}_{\text{cen}} = \frac{1}{\sum_{i=1}^N \mathbbm{1}(\vect{p}_i)}\sum_{i=1}^N \mathbbm{1}(\vect{p}_i) \, \left|h_i - \exp\left(-\frac{\gamma{\|\vect{x}_i\|}^2}{b_i^2}\right)\right|,
\end{equation}
with $b_i$ denoting the length of the axis-aligned box, $\gamma$ the temperature hyperparameter, and $\mathbbm{1}(\vect{p}_i)$ an indicator function that returns 1 if and only if $\vect{p}_i$ belongs to an instance. We then generate candidates from predicted heatmaps via local normalized non-maximum suppression (LN-NMS), with duplicate proposals being aggregated based on their context. The aggregation is supervised via a ground-truth merging map $\hat a$:
\begin{equation}
    \mathcal{L}_{\text{agg}} = H_b(a, \hat a),
\end{equation}
with $H_b$ denoting the binary cross-entropy loss.

 An MLP processes $\vect{f}_{\text{3D}}$ to produce pointwise instance features $\vect{g} \in \mathbb{R}^{N\times{}L}$.
 The generated candidate masks are then used to average-pool instance features $\vect{g}$ across each mask in order to generate the instance embeddings $\vect{e_i} \in \mathbb{R}^{I\times{}L}$ that are input to the subsequent verbo-visual fusion module, where $I$ is the number of candidates. For further details regarding the heatmap generation and proposal aggregation, we refer the readers to Wu~\etal~\cite{wu2022dk}.

The resulting total loss to supervise candidate generation is:
\begin{equation}
    \mathcal{L}_{\text{can}} = \mathcal{L}_{\text{sem}} + \mathcal{L}_{\text{off}} + \mathcal{L}_{\text{cen}} + \mathcal{L}_{\text{agg}}.
\end{equation}

\noindent\textbf{Mask generation.} To generate dense instance masks, we first remap each instance candidate onto its respective kernel parameters via an MLP, and following DyCo3d~\cite{he2021dyco3d}, we generate the final instance masks $\vect{z}\in\{0,1\}^N$ with the use of dynamic convolutions. The masks that have ground-truth counterparts (IoU $\geq$ 0.25) can then be supervised via:
\begin{equation}
    \mathcal{L}_{\text{mask}} = H_b(\vect{z}, \hat{\vect{z}}) + \text{DICE}(\vect{z}, \hat{\vect{z}}),
\end{equation}
with DICE denoting the Dice loss.

\subsection{Encoding Language Cues}
\label{sec:language}

In recent years, NLP has boomed with the success of large pre-trained transformer models that perform exceptionally well on a wide range of different tasks~\cite{devlin2018bert, liu2019roberta, song2020mpnet}. Such models are trained with large-scale datasets that allow them to capture context and intent within natural language prompts. While most early work in 3D visual grounding employed the more traditional GloVE embeddings~\cite{pennington2014glove} followed by a GRU~\cite{chen2020scanrefer, zhao20213dvgt, chen2022ham}, recent works have switched focus towards transformer-based verbal encoding strategies~\cite{jain2022bottom, roh2022languagerefer}.
In this work, to extract the initial word embeddings, we utilize a pre-trained transformer architecture, namely MPNet~\cite{song2020mpnet}. We then project the encoded tokens to $L$ dimensions via by a single linear layer to form the word embeddings $\vect{e_w} \in \mathbb{R}^{W \times L}$, with $W$ denoting the number of tokens. The generated word embeddings are then used as input for the subsequent verbo-visual fusion module.

\subsection{Verbo-visual Fusion}
\label{sec:grounding}

Commonly, 3D visual grounding is achieved through grounding-by-selection, wherein given a set of visual candidates, a verbal cue is used to select the referred object~\cite{chen2020scanrefer, zhao20213dvgt, chen2022ham}. Formally, a fusion module consumes both visual and verbal information to output a probability distribution over predetermined candidates in order to determine the likelihood that an object is referred by the description, the argmax of which is taken as the prediction. While early approaches have considered a simple MLP to fuse the two modalities~\cite{chen2020scanrefer}, recent methods are constructed using the popular transformer architecture~\cite{zhao20213dvgt, chen2022ham}, taking advantage of the expressibility of the attention mechanism given by:
\begin{equation} \label{eq:attention}
    \vect{f}_l = \textrm{softmax}(\vect{q}_l \vect{k}_l^T)\vect{v}_l + \vect{f}_{l-1},
\end{equation}
with the queries $\vect{q}$ extracted from object features and key-value pairs $\vect{k}$ and $\vect{v}$ interchangeably from object and word features.

The naive approach when tackling 3D visual grounding in a \textit{dense} setting is to follow a similar pipeline. Given instance candidates $\vect{e_i}$ (Sec.~\ref{sec:instance}) and word embeddings $\vect{e_w}$ (Sec.~\ref{sec:language}), a transformer decoder fuses the multi-modal information and ultimately selects the referred mask. However, while instance segmentation has shown to perform on par, if not better than 3D object detection for 3D indoor localization\footnote{Classifying each point yields a more robust solution compared to the localization of 8 corner points in complete free 3D space.}, dense kernel-based models show limited separability in the latent space between instances of the same semantic class. When it comes to 3DVG, this results in a significant performance drop when facing utterances that refer to repetitive instances, i.e., instances that are not semantically unique in a scene. To combat this, we propose four modules that (i) aim to disambiguate inter-instance relational cues, (ii) aid training to induce better separability in the latent representation, (iii) infer the sensor position to resolve view-dependent descriptions, {and (iv) ensemble across multiple viewpoints to improve mask quality}.

\noindent \textbf{Bottom-up attentive fusion (BAF).} Natural language prompts may aim to localize a repetitive object by establishing its relation to another object (e.g. ``The chair next to the cabinet.''). We observe that due to human nature, as our attention is limited, verbal relations are often formed between neighboring instances. However, such localized information routing is difficult to learn for global attention schemes. Inspired by the effective windowing in NLP and 2D applications~\cite{zhang2021longformer,cheng2022mask2former}, to explicitly induce this locality condition in our model, we develop a bottom-up attentive fusion module (BAF) built on a masked self-attention mechanism. An illustration of BAF can be seen in Fig.~\ref{fig:pipeline} - right.

In BAF, the word embeddings first get processed via a vanilla transformer encoder block for further abstraction. Next, the spatial and contextual information between candidate instances are routed via a localized self-attention layer. Formally, we restate the mechanism of Eq.~\ref{eq:attention} for the localized self-attention layer:
\begin{equation}
    \vect{f}_l = \textrm{softmax}(\vect{M}_l + \vect{q}_l \vect{k}_l^T)\vect{v}_l + \vect{f}_{l-1}.
\end{equation}
The mask takes the value $\vect{M}_l(i,j)$ when computing the attention vector between the $i$'th and $j$'th instance candidates:
\begin{equation} \label{eq:masking}
\vect{M}_l(i,j) = \begin{cases}
0, & \textrm{if } || \vect{o}_i - \vect{o}_j || < r_l \\
-\infty,\quad & \textrm{otherwise}
\end{cases}
\end{equation}
with $r_l$ denoting the radius of the spherical masking set per layer. The spherical 3D masking operation limits the attention to neighboring instances, which helps when grounding cues contain inter-instance relations.

Via a cross-attention and feed-forward layer, we fuse the instance tokens with word embeddings. To construct the final instance embeddings, we apply the transformer decoder block $l$ times, each time with an increasing masking radius $r_l$ (hence bottom-up), and fuse the resulting features across different layers via an MLP to capture relational cues from different neighborhood scales. We attach a classification head that maps the language-aware instance tokens onto confidence scores $\vect{u}\in\mathbb{R}^{I}$. We supervise the selection via cross-entropy:
\begin{equation}
    \mathcal{L}_{sel} = H(\vect{u}, \hat{\vect{u}}),
\end{equation}
with $\hat{\vect{u}}$ denoting the ground-truth index computed by applying the Hungarian algorithm on the instance predictions and referred ground-truth mask. The final dense visual grounding prediction is then obtained via:
\begin{equation}
    \vect{z}^* = \vect{z}^{(i)} \, | \, i = \argmax \vect{u}.
\end{equation}

\noindent\textbf{Inducing separation via contrastive learning.}
With the aim of alleviating ambiguities between repetitive instance mappings, we propose employing a verbo-visual contrastive learning scheme to induce better separation between language embeddings and language-aware instance embeddings in the latent space. Firstly, we form sentence embeddings by applying masked averaging to the learned word embeddings (${\vect{e_s}=\overline{\vect{e_w}}}$). In the contrastive loss formulation, matched sentence embeddings and instance vectors are treated as positive samples and thus are pulled towards each other, while the pairings with remaining instances are treated as negative samples and are pushed away from one another. Formally, this can be expressed as:
\begin{equation}
    \mathcal{L}_{\text{con}}(\vect{e_s}, \vect{e_i}) = - \log \frac{\exp(d(\vect{e_s}, \vect{e_i}_{,k+}) / \tau)}{\sum_{k} \exp(d(\vect{e_s}, \vect{e_i}_{,k}) / \tau)},
\end{equation}
with $\tau$ denoting the temperature, $\vect{e_s}$ and $\vect{e_i}_{,k}$ referring to the sentence embedding and $k$'th instance candidate embeddings respectively. $d(\cdot)$ is chosen as the cosine similarity and $k+$ denotes the index of the instance that matches the reference cue. The contrastive loss provides an easy-to-apply, general solution to aid referral-based localization within the \textit{multiple} subset, i.e., the subset of repetitive instance referrals.

\noindent \textbf{Global camera token (GCT).} As individuals, we observe the world from our own personal perspective. It is often this perspective that we tap into to describe our surroundings, which results in view-dependent references that may become impossible to decipher or disambiguate in 3D space. To combat, we propose learning the camera position as an auxiliary task.

We input a learned camera embedding into the bottom-up attentive fusion module alongside the instance candidates. The camera embedding is treated as a global token, i.e.\ all instances regardless of centroid position are allowed to attend and be attended by the camera token on all layers. Formally, we restate the masking value from Eq.~\ref{eq:masking} for the given camera token index $i_c$:
\begin{equation}
    \vect{M}_l(i,i_c) =  \vect{M}_l(i_c,i) = 0, \ \forall \ i.
\end{equation}
We supervise the output global camera token $\vect{t}$ with the camera positions that were used during the annotation process $\hat{\vect{t}}$:
\begin{equation}
    L_2(\vect{t}, \hat{\vect{t}}),
\end{equation}
with $L_2$ denoting the L2 loss.

\begin{wrapfigure}{r}{0.5\textwidth}
\vspace{-3px} \begin{minipage}{0.5\textwidth}
    \begin{algorithm}[H]
    \SetKwInput{KwInput}{Input}
    \SetKwInput{KwOutput}{Return}
    \DontPrintSemicolon
    \KwInput{$P$, $D$, $\theta$, $R$, $\tau$}
    \For{$r$ in $R$} {
        P$_r$ = rotate(P, $r$) \;
        pred = $\theta$(P$_r$, D) \ $\in \{0,1\}^N$ \;
        preds.append(pred) \;
    }
    energy = pairwise\_iou(preds) \;
    seed = energy.sum(1).argmax() \;
    preds = preds[energy[seed] $> \tau$] \;
    pred = preds.sum(0) $>$ (num\_preds / 2) \;
    \KwOutput{pred $\in \{0,1\}^N$}
    \caption{MVE}
    \label{alg:tta}
    \end{algorithm}
\end{minipage}
\vspace{5pt} \end{wrapfigure}

\noindent {\textbf{Multi-view Ensembling (MVE).}
Point clouds are irregular data structures that retain their order given an affine transformation, e.g. the order of the points does not change under rotation. This property is commonly exploited to generate a multi-view representation of a given scene while still retaining point-to-point correspondences~\cite{huang2022multiview, guo2023viewrefer}. As a final improvement, we propose a multi-view ensembling approach that leverages this property to improve the quality of the predicted referred masks.}

{
From a point cloud $P$, we construct $K$ inputs $P_r \in \mathbb{R}^{N \times 6}$, each rotated with a different yaw rotation $r \in R = [0, ..., 2\pi) \in \mathbb{R}^{K}$. Given a natural language description $D$, we predict the dense visual grounding mask for each input pair ($P_r$, $D$) (Alg.~\ref{alg:tta}~L1-4). Due to the variations in the input point cloud under different yaw rotations, the predicted referred masks may vary in two ways: (i) the selection might vary, i.e. not all forward passes may generate the instance mask of the same object (ii) the segmentation results may vary, i.e. even if the same object is selected, the set of points defining the mask might be different. With MVE, we tackle the two issues consecutively by first determining the most likely target object and only then refining the mask prediction.
}

{
 In our two-step approach, (Step 1) we start by computing the pairwise IoUs of all $K$ predictions to form an energy matrix $E \in [0,1]^{K \times K}$ (Alg.~\ref{alg:tta}~L5). We determine the seed mask as the predicted mask that shows the highest IoU to all other predictions (Alg.~\ref{alg:tta}~L6). (Step 2) Once the seed mask is determined, we accept masks as valid if they aim to localize the same object, i.e. if the IoU to the seed mask exceeds a predetermined threshold $\tau$ (Alg.~\ref{alg:tta}~L7). In the second step, we compute the final prediction via pointwise majority voting amongst the valid predictions (Alg.~\ref{alg:tta}~L8). Our proposed improvement aims to reduce the epistemic uncertainty within not only the selection process (Step 1 for visual grounding) but also the mask prediction (Step 2 for instance segmentation).
}

\section{Experiments}

\begin{table}[t]
    \tabcolsep=0.015cm
    \centering
    \caption{Comparison to state-of-the-art 3D visual grounding methods evaluated on the ScanRefer \textit{val}-set as well as its online \textit{test} server. Reported are the accuracy values at $25\%$/$50\%$ IoU thresholds, with the main metric being the overall accuracy at $50\%$ threshold (Acc@50). TTA: test-time augmentation. ConcreteNet not only outperforms existing work, but also predicts dense 3D masks with the potential of aiding higher-level tasks that require a finer geometric understanding of an instance.} \vspace{-5px}
    \begin{tabular}{|c|l|cc|cc|cc|cc|}
         \cline{2-10}
         \multicolumn{1}{c|}{} & & & & \multicolumn{2}{c|}{Unique} & \multicolumn{2}{c|}{Multiple} & \multicolumn{2}{c|}{Overall} \\
         \multicolumn{1}{c|}{} & Method  & Input & Output & Acc@25 & Acc@50 & Acc@25 & Acc@50 & Acc@25 & \textbf{Acc@50} \\
         \hline
         \parbox[t]{2mm}{\multirow{13}{*}{\rotatebox[origin=c]{90}{\textit{val}-split}}} & ScanRefer~\cite{chen2020scanrefer}  & 3D & Box & 67.64 & 46.19 & 32.06 & 21.26 & 38.97 & 26.10 \\
         & TGNN~\cite{huang2021tgnn}  & 3D & Mask & 68.61 & 56.80 & 29.84 & 23.18 & 37.37 & 29.70 \\
         & SAT~\cite{yang2021sat}  & 2D+3D & Box & 73.21 & 50.83 & 37.64 & 25.16 & 44.54 & 30.14 \\
         & InstanceRefer~\cite{yuan2021instancerefer}  & 3D & Mask & 77.45 & 66.83 & 31.27 & 24.77 & 40.23 & 32.93 \\
         & 3DVG-T~\cite{zhao20213dvgt}  & 3D & Box & 77.16 & 58.47 & 38.38 & 28.70 & 45.90 & 34.47 \\
         & MVT~\cite{huang2022multiview}  & 2D+3D & Box & 77.67 & 66.45 & 31.92 & 25.26 & 40.80 & 33.26 \\
         & 3DJCG~\cite{cai20223djcg}  & 3D & Box & 78.75 & 61.30 & 40.13 & 30.08 & 47.62 & 36.14 \\
         & D3Net~\cite{chen2022d3net}  & 2D+3D & Box & - & 70.35 & - & 30.05 & - & 37.82 \\
         & BUTD-DETR~\cite{jain2022bottom}  & 3D & Box & - & - & - & - & 52.20 & 39.80 \\
         & HAM~\cite{chen2022ham}  & 3D & Box & 79.24 & 67.86 & 41.46 & 34.03 & 48.79 & 40.60 \\
         & EDA~\cite{wu2023eda} & 3D & Box & 85.76 & 68.57 & \textbf{49.13} & 37.64 & \textbf{54.59} & 42.26 \\
         & M3DRef-CLIP~\cite{zhang2023multi3drefer} & 3D & Box & - & 77.2\phantom{0} & - &  36.8\phantom{0} & - & 44.7\phantom{0} \\
         \cline{2-10}
         & ConcreteNet & 3D & Mask & \textbf{86.40} & \textbf{82.05} & 42.41 & \textbf{38.39} & 50.61 & \textbf{46.53} \\
         \hline \hline
         \parbox[t]{2mm}{\multirow{10}{*}{\rotatebox[origin=c]{90}{\textit{test}-split}}} & ScanRefer~\cite{chen2020scanrefer}  & 2D+3D & Box & 68.59 & 43.53 & 34.88 & 20.97 & 42.44 & 26.03 \\
         & TGNN~\cite{huang2021tgnn}  & 3D & Mask & 68.34 & 58.94 & 33.12 & 25.26 & 41.02 & 32.81 \\
         & InstanceRefer~\cite{yuan2021instancerefer}  & 3D & Mask & 77.82 & 66.69 & 34.57 & 26.88 & 44.27 & 35.80 \\
         & 3DVG-T~\cite{zhao20213dvgt}  & 2D+3D & Box & 77.33 & 57.87 & 43.70 & 31.02 & 51.24 & 37.04 \\
         & 3DJCG~\cite{cai20223djcg}  & 2D+3D & Box & 76.75 & 60.59 & 43.89 & 31.17 & 51.26 & 37.76 \\
         & D3Net~\cite{chen2022d3net}  & 2D+3D & Box & 79.23 & 68.43 & 39.05 & 30.74 & 48.06 & 39.19 \\
         & BUTD-DETR~\cite{jain2022bottom}  & 3D & Box & 78.48 & 54.99 & 39.34 & 24.80 & 48.11 & 31.57 \\
         & HAM~\cite{chen2022ham}  & 3D & Box & 77.99 & 63.73 & 41.48 & 33.24 & 49.67 & 40.07 \\
         & M3DRef-CLIP~\cite{zhang2023multi3drefer} & 3D & Box & 79.80 & 70.85 & 46.92 & 38.07 & 54.33 & 45.45 \\
         \cline{2-10}
         & ConcreteNet & 3D & Mask & \textbf{86.07} & \textbf{79.23} & \textbf{47.46} & \textbf{40.91} & \textbf{56.12} & \textbf{49.50} \\
         \hline
    \end{tabular}
    \label{tab:results_test}
\vspace{-10px} \end{table}

We carry out our experiments and extensively ablate our components on the ScanRefer dataset~\cite{chen2020scanrefer}, which provides 51,583 descriptions of 11,046 objects from 800 ScanNet scenes~\cite{dai2017scannet}. To remain comparable to the existing literature and follow the evaluation guidelines set by the dataset, we fit an axis-aligned bounding box onto our predicted instance masks and evaluate our method using these bounding boxes. We report the accuracy [$\%$] at IoU thresholds of $25\%$ and $50\%$, further providing a split between \textit{unique} and \textit{multiple} subsets, with \textit{unique} referring to instances that have a unique semantic class in a given scene.

In the supplement, we further provide the implementation details along with the evaluation for referral-based 3D object localization on the Nr3D dataset~\cite{achlioptas2020referit3d}.


\subsection{Results}

\begin{table}[t]
    \centering
    \caption{Ablation study on each proposed component on the ScanRefer \textit{val}-set. Starting with 3DVG-T with a 3D instance segmentation backbone, we first replace the word encoder (NLP BB), then systematically introduce our proposed bottom-up attentive fusion module (BAF), contrastive loss ($\mathcal{L}_{con}$), and finally the learned global camera token (GCT) and {the multi-view ensembling (MVE).}}\vspace{-5px} 
    \begin{tabular}{|ccccc|ccc|}
    \hline
    & & & & & \multicolumn{3}{c|}{Acc@50} \\
    NLP BB & BAF & $\mathcal{L}_{con}$ & GCT & MVE & Unique & Multiple & Overall \\
    \hline
    GloVe & & & & & 75.73 & 27.64 & 36.60 \\
    MPNet & & & & & 76.64 & 27.86 & 36.95 \\
    MPNet & \ding{51} & & & & 74.49 & 33.43 & 41.08 \\
    MPNet & \ding{51} & \ding{51} & & & 76.47 & 34.00 & 41.91 \\
    MPNet & \ding{51} & \ding{51} & \ding{51} & & 75.62 & 36.56 & 43.84 \\
    MPNet & \ding{51} & \ding{51} & \ding{51} & \ding{51} & \textbf{82.05} & \textbf{38.39} & \textbf{46.53} \\
    \hline
    \end{tabular}
    \label{tab:ablation_component}
\vspace{-5px} \end{table}

In Tab.~\ref{tab:results_test} we report the performance of our proposed ConcreteNet on the ScanRefer \textit{val}-set as well as the online $test$-server and compare it to existing methods. As seen, ConcreteNet significantly outperforms existing work in both unique and \textit{multiple} subsets {at the more difficult $50\%$ IoU threshold}, while also outputting dense 3D instance masks---as opposed to 3D bounding boxes---to aid higher-level tasks that require physical interactions and fine geometric detail.

\subsection{Ablation Studies}

\begin{figure}[t]
    \centering
    \includegraphics[width=\linewidth]{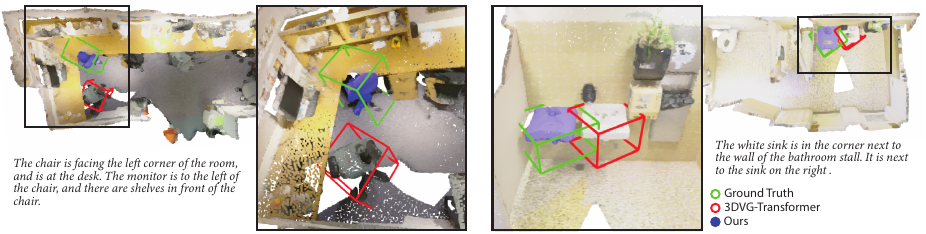}
    \caption{Qualitative results from the ScanRefer \textit{val}-set depicting the dense predictions of ConcreteNet against the ground truth and 3DVG-Transformer~\cite{zhao20213dvgt} predictions. We showcase two cases that illustrate the effectiveness of BAF (left) and GCT (right).}
    \label{fig:results}
 \vspace{-15px} \end{figure}

\begin{table}[t]
\centering
\begin{minipage}{.45\textwidth}
    \centering
     \caption{Comparing a 3D detector vs. segmenter as the visual backbone.} \vspace{-5px}
    \begin{tabular}{|lc|cc|}
    \hline
    & & \multicolumn{2}{c|}{Overall} \\
    Method & Output & Acc@25 & Acc@50 \\
    \hline
    \multirow{2}{*}{ScanRefer~\cite{chen2020scanrefer}} & Box & \textbf{38.97} & 26.10 \\
    & Mask & 33.18 & \textbf{28.69} \\
    \hline
    \multirow{2}{*}{3DVG-T~\cite{zhao20213dvgt}} & Box & \textbf{45.90} & 34.47 \\
    & Mask & 42.29 & \textbf{36.60} \\
    \hline
    \end{tabular}
    \label{tab:ablation_visual_backbone}
\end{minipage}
\hfill
\hfill
\begin{minipage}{.53\textwidth}
    \centering
    \caption{Ablation study on the attentive fusion module comparing global attention with a masking approach.} \vspace{-5px}
    \begin{tabular}{|l|ccc|}
    \hline
    & \multicolumn{3}{c|}{Acc@50} \\
    Attention & Unique & Multiple & Overall \\
    \hline
    Global (Baseline) & 76.64 & 27.86 & 36.95 \\
    Top-Down & \textbf{77.77} & 32.16 & 40.66 \\
    Bottom-Up (Ours) & 74.49 & \textbf{33.43} & \textbf{41.08} \\
    \hline
    \end{tabular}
    \label{tab:ablation_baf}
\end{minipage}
 \vspace{-5px} \end{table}

\begin{table}[t]
    \centering
    \caption{Comparison of our baseline ConcreteNet (i) \textbf{without} using any camera information (ii) via a \textbf{learned} a global camera token (GCT) (iii) by using the camera position as direct \textbf{input}.} \vspace{-5px}
     \begin{tabular}{|ll|cc|cc|cc|}
         \hline
         & & \multicolumn{2}{c|}{Unique} & \multicolumn{2}{c|}{Multiple} & \multicolumn{2}{c|}{Overall} \\
         \multicolumn{2}{|l|}{Camera} & Acc@25 & Acc@50 & Acc@25 & Acc@50 & Acc@25 & \textbf{Acc@50} \\
         \hline
         \multicolumn{2}{|l|}{Without} & 83.58 & 76.47 & 39.05 & 34.00 & 47.35 & 41.91 \\
         \multirow{2}{*}{GCT} & Learned & 82.39 & 75.62 & 41.24 & 36.56 & 48.91 & 43.84   \\
         & Input & \textbf{88.15} & \textbf{79.80} & \textbf{51.42} & \textbf{44.62} & \textbf{58.27} & \textbf{51.18} \\
         \hline
        \end{tabular}
        \label{tab:ablation_camera}
 \vspace{-10px} \end{table}

\noindent \textbf{Effects of individual components.} In Tab.~\ref{tab:ablation_component} we showcase an ablation study where we isolate the effects of our proposed components. Starting
with the reimplementation of 3DVG-T~\cite{zhao20213dvgt} with our 3D instance segmentation backbone, we first replace the GloVe+GRU~\cite{pennington2014glove} verbal encoder with the more recent transformer-based MPNet~\cite{song2020mpnet} for a minor boost to performance across the board. Next, we introduce the bottom-up attentive fusion module (BAF) and the contrastive loss $\mathcal{L}_{\text{con}}$ that aim to disambiguate repetitive instance embeddings. As expected, we observe a significant boost in the overall accuracy of our model stemming purely from the \textit{multiple} subset. {Finally, we include the global camera token (GCT), and our multi-view ensembling (MVE) which combined allow ConcreteNet to reach $46.53\%$ accuracy with the $50\%$ IoU threshold.}

We further show qualitative results from the ScanRefer \textit{val}-set that demonstrate the effects achieved by our proposed components. As seen in Fig.~\ref{fig:results} - left, while the prompt may be construed as valid for both chairs on the left of the room, the neighborhood masking introduced via BAF allows ConcreteNet to correctly identify the referred chair. Furthermore, in  Fig.~\ref{fig:results} - right, we demonstrate the necessity of the learned GCT, where under a view-dependent prompt, the referred sink is accurately segmented.

\noindent \textbf{Segmentation vs.\ detection for grounding.} To illustrate the effects of the visual backbone change (from detection to segmentation), we conduct an ablation study where we replace the 3D object detection backbones of existing methods with our kernel-based 3D instance segmentation backbone based on DKNet~\cite{wu2022dk}. As seen in Tab.~\ref{tab:ablation_visual_backbone}, dense 3D masks yield more stable results across the two IoU thresholds, resulting in a better performance for accuracy at $50\%$ IoU and lower performance at $25\%$. In the supplementary materials, we also provide the detailed results for the \textit{unique} and \textit{multiple} subsets, where it can be seen that 3D instance segmentation yields significantly better accuracy for unique prompts, yet performs much worse in the \textit{multiple} subset. While 3D instance segmentation has the benefit of robustness in localization, the reduced performance for the \textit{multiple} subset shows that its kernel features lack enough separation in the latent space to effectively disambiguate repetitive instances.


\noindent \textbf{Bottom-up attention masking.} We introduce spherical attentive masking in a bottom-up manner to aid grounding referrals that may be construed as valid for multiple instances. In Tab.~\ref{tab:ablation_baf} we showcase the necessity of this masking operation by comparing to a purely global attention strategy. While our bottom-up method performs on-par when grounding unique instances, as expected we observe a substantial boost in the \textit{multiple} subset ($+4.13\%$). Furthermore, in Tab.~\ref{tab:ablation_baf} we also compare the bottom-up strategy to a top-down approach. Here we observe that while spherical masking results in improvements in the \textit{multiple} subset for both cases, inducing locality at early stages does further help with the disambiguation (+1.27\%).

 
\noindent \textbf{Learning camera information.} In Tab.~\ref{tab:ablation_camera} we show that learning a global camera token results with major improvements across the board. We speculate that while the major direct benefits of a learned camera token come from the \textit{multiple} subset, the reduced ambiguity in view-dependent prompts further reduces the overall dataset noise, allowing better use of available capacity, which also benefits the unique cases. This is further seen when instead of a learned GCT, we directly input encoded camera information. As seen in Tab.~\ref{tab:ablation_camera}, directly inputting camera information yields an unprecedented improvement over the baseline method with accuracy at $50\%$ IoU reaching $51.18\%$\footnote{We believe that input camera positions are a reasonable assumption in indoor robotic applications and hope that this performance potential will motivate future research.}. While we showcase the potential benefits of such input information, its realization in a learned setting is not trivial due to the ill-posed nature of determining camera positions from an unlabeled and noisy dataset, as most prompts may not contain any view-dependent clues or those that do might yield a wide range of feasible solutions.

\section{Conclusion}

In this work, we tackle the problem of \textit{dense} 3D visual grounding, i.e.\ referral-based 3D instance segmentation. We establish a baseline kernel-based dense 3D grounding approach and tackle its arisen weaknesses by proposing four standalone improvements. We introduce a bottom-up attentive fusion module to localize inter-instance relational cues, construct a contrastive loss to induce latent space separation, learn a global camera token to disambiguate view-dependent utterances, {and finally ensemble multiple viewpoints to refine the referred prediction}. Combining these four modules, our proposed ConcreteNet sets the new state of the art on the popular ScanRefer online benchmark.

\vspace{5px}
\noindent {\textbf{Limitations:} We discuss the limitations of ConcreteNet in the supplement.}

\vspace{5px}
\noindent\textbf{Acknowledgments:} This work is funded by Toyota Motor Europe via the research project TRACE-Z\"urich.
\clearpage
\clearpage

\bibliographystyle{splncs04}
\bibliography{main}


\clearpage
\clearpage

\appendix

\section{Implementation Details}

The visual inputs are formed via the concatenation of 3D coordinates and RGB color channels. For the 3D Unet backbone we use a voxel size of $2cm$ following set convention~\cite{jiang2020pointgroup, vu2022softgroup, wu2022dk}. For the natural language encoding, we use the MPNet tokenizer and pre-trained model~\cite{song2020mpnet, wolf2020transformers}. Following Zhao~\etal~\cite{zhao20213dvgt}, we randomly mask the referred object nouns with a probability of $0.5$ before the extraction of word embeddings in order to reduce overfitting and entice learning context to aid localization. We remap the output of MPNet onto a $d=128$ dimensional vector using a single linear layer. BAF is built using a vanilla transformer encoder (2-layer) for the language encoding and a decoder (6-layer) for the attentive fusion~\cite{vaswani2017attention}. After every other decoder layer, we increase the radius of the masking sphere $r_l$ from [$1.0m$, $2.5m$, $\infty$], with $2.5m$ giving the approximate average inter-instance distance and $\infty$ providing global attention in the final two layers. We set $\gamma = 25$ following DKNet~\cite{wu2022dk}, and $\tau = 0.3$. We empirically choose $K=5$ and $\tau=0.9$ for MVE. We use a batch size of 4, with each sample consisting of a single scene and up to 32 utterances. We train for 400 epochs using the AdamW optimizer~\cite{loshchilov2017decoupled} with a learning rate of $3\cdot10^{-4}$ using a single Nvidia RTX 3090.

\section{Segmentation vs. Detection for Grounding}

In Tab.~\ref{tab:ablation_visual_backbone_extended} we provide an extended analysis of the visual backbone change for 3D grounding (main paper Tab.~3).

\begin{table}[t]
    \centering
    \caption{Extension of Tab.~3: We replace the 3D object detector (outputs bounding box) of established 3D visual grounding models with our 3D instance segmentation backbone (outputs mask) to showcase the performance implications.} \vspace{-5px}
    \begin{tabular}{|cc|cc|cc|cc|}
         \hline
         & & \multicolumn{2}{c|}{Unique} & \multicolumn{2}{c|}{Multiple} & \multicolumn{2}{c|}{Overall}  \\
         Method & Output & Acc@25 & Acc@50 & Acc@25 & Acc@50 & Acc@25 & Acc@50 \\
         \hline
         \multirow{2}{*}{ScanRefer} & Box & 67.64 & 46.19 & \textbf{32.06} & \textbf{21.26} & \textbf{38.97} & 26.10 \\
          & Mask & \textbf{70.49}  & \textbf{63.26} & 24.64 & 20.77 & 33.18 & \textbf{28.69} \\
          \hline
          \multirow{2}{*}{3DVG-T} & Box & 77.16 & 58.47 & \textbf{38.38} & \textbf{28.70} & \textbf{45.90} & 34.37 \\
           & Mask & \textbf{82.34} & \textbf{75.73} & 33.12 & 27.64 & 42.29 & \textbf{36.60} \\
          \hline
    \end{tabular}
    \label{tab:ablation_visual_backbone_extended}
\end{table}

\begin{table}[t]
    \tabcolsep=0.185cm
    \centering
    \caption{Comparison of SOTA on Nr3D evaluated on referral-based 3D object localization a la ScanRefer.} \vspace{-5px}
    \begin{tabular}{|l|ccccc|}
    \hline
    Method & SRefer~\cite{chen2020scanrefer} & 3DVG-T~\cite{zhao20213dvgt} & D3Net~\cite{chen2022d3net} & HAM~\cite{chen2022ham} & Ours \\
    \hline
    Acc@50 & 12.17 & 14.22 & 25.23 & 27.11 & \textbf{33.66} \\
    \hline
    \end{tabular}
    \label{tab:nr3d}
\end{table}

\begin{table}[t]
    \tabcolsep=0.29cm
    \centering
    \caption{Evaluating 3D dense visual grounding on the ScanRefer \textit{val}-set, where the IoU is determined based on not the bounding boxes but the instance masks. Shown are the overall Acc@25/50.} \vspace{-5px}
    \begin{tabular}{|c|c|c|c|c|}
    \hline
    Baseline & +BAF & +$\mathcal{L}_{con}$ & +GCT & +MVE \\
    \hline
    44.2 / 39.1 & 46.9 / 43.3 & 49.2 / 44.9 & 50.7 / 46.8 & 51.4 / 48.6\\
    \hline
    \end{tabular}
    \label{tab:dense_eval}
\end{table}

\section{NR3D}

While ScanRefer tackles referral-based object localization, Nr3D~\cite{achlioptas2020referit3d} is built on referral-based identification. This means that in Nr3D, the ground-truth bounding boxes/instance masks are assumed to be given as input. We instead tackle the more challenging task of end-to-end referral-based object localization. Even though our method is not directly applicable to the Nr3D benchmark, the dataset can still be used to evaluate our method for our task of interest.

As seen in Tab.~\ref{tab:nr3d}, ConcreteNet significantly outperforms existing methods on Nr3D by $+6.55\%$ overall accuracy at $50\%$ threshold, despite not being able to utilize the global camera token due to a lack of available camera information.

\section{Evaluating 3D \textit{Dense} Visual Grounding}
In Tab.~\ref{tab:dense_eval} we repeat the component-wise ablation study (see Tab.~2 of the main paper), but with the IoU computed on the instance masks rather than the axis-aligned bounding boxes. Here we again observe that each component significantly improves the \textit{dense} 3D grounding performance as well.

Specifically, we see an improvement of $+4.2\%$, $+1.6\%$, $+1.9\%$ and $+1.8\%$ in accuracy at the $50\%$ threshold when one-by-one introducing the bottom-up attentive fusion module (BAF), the contrastive loss ($\mathcal{L}_{con}$), the global camera token (GCT) and the multi-view ensembling (MVE) respectively.

\begin{table}[t]
    \caption{Ablation study on GCT reporting Acc@50.} \vspace{-5px}
    \tabcolsep=0.277cm
    \centering
    \begin{tabular}{|l|ccc|}
    \hline
    Method & Unique & Multiple & Overall \\
    \hline
    GCT (proposed) & 75.62 & \textbf{36.56} & \textbf{43.84} \\
    GCT (with pose) & \textbf{76.35} & 35.21 & 42.88 \\
    \hline
    \end{tabular}
    \label{tab:rebuttal_gct}
\end{table}

\section{Camera Rotation in GCT}

We have only include camera \emph{position} in GCT as it is sufficient for disambiguating view-dependent prompts (e.g. left, right relations). Tab.~\ref{tab:rebuttal_gct} shows a further comparison that extends GCT to include yaw and pitch. We supervise the unit look-at direction vector via a cosine similarity loss to the ground truth. This addition of rotational information slightly hurts the overall performance, especially for \textit{multiple} cases. We hypothesize that adding these uninformative (as justified above) rotational dimensions to the GCT target merely increases problem complexity and does not help disambiguation.

\section{Additional Qualitative Results}

\begin{figure}[t]
    \centering
    \includegraphics[width=0.49\textwidth]{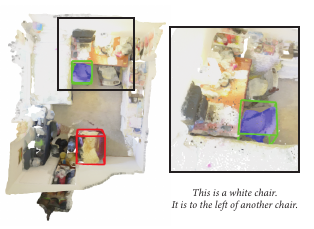}
    \caption{Additional qualitative result from the ScanRefer \textit{val}-set showing the benefits of a learned global camera token.}
    \label{fig:results_supp2}
\end{figure}

In Fig.~\ref{fig:results_supp2} further qualitative results from the ScanRefer \textit{val}-set can be seen that demonstrate the benefits of the global camera token (GCT).

\begin{figure}[t]
    \centering
    \includegraphics[width=\textwidth]{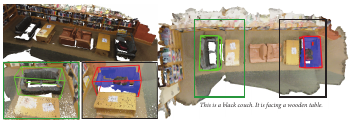}
    \caption{Failure case from the ScanRefer \textit{val}-set. While the predictions from both ConcreteNet and 3DVG-Transformer do not match the ground truth, given the symmetric nature of the scene along with the vagueness of the description, it can be seen that the cue does match both predictions and the ground truth.}
    \label{fig:results_supp1}
\end{figure}

Additionally, in Fig.~\ref{fig:results_supp1} we show a common failure case where, while the model predictions do not match the ground-truth object, the natural language description still fits the output.

\section{Analysing the Semantic Class Accuracy of the Model Predictions}

An analysis of the predicted instance semantics can be found in Tab.~\ref{tab:semantic_class_accuracy}. Specifically, we extract the semantic label of the predicted referral-based instance mask from the ground-truth semantic labels. We then report the accuracy when comparing the predicted semantic class to the ground-truth counterpart. It can be seen that with $\sim86\%$ accuracy, ConcreteNet is able to correctly identify the semantic class of the referred object instance in most cases. Furthermore, from the marginal gap between the \textit{unique}  semantic class accuracy and the \textit{unique} accuracy at $25\%$ IoU (Learned GCT on Tab.~5 of the main paper), the effectiveness and robustness of the 3D visual backbone can be inferred.

\begin{table}[t]
    \centering
    \caption{Further analysis on the semantic class of the predicted instance without the inclusion of MVE. We extract the semantic class of the predicted mask and measure the accuracy compared to the semantic class of the target object instance.} \vspace{-5px}
    \tabcolsep=0.29cm
    \begin{tabular}{|ccc|}
        \hline
         Unique & Multiple & Overall \\
        \hline
         85.33 & 86.22 & 86.05  \\
        \hline
    \end{tabular}
    \label{tab:semantic_class_accuracy}
\end{table}

\section{Limitations and Discussion}

In this section, we dive into the limitations of our work and discuss possible remedies and counterpoints.

\noindent \textbf{Global Camera Token}: To employ a learned global camera token (GCT), ConcreteNet requires ground-truth camera information to be provided. As this can be trivially collected during a standard data annotation pipeline, we hope that the presented benefits of GCT aid in conveying the necessity of such additions in future 3D visual grounding datasets. Nevertheless, as seen in Tab.~\ref{tab:nr3d}, our method continues to show state-of-the-art performance when trained and evaluated without the inclusion of GCT on datasets that lack such information.

\sloppy{\noindent \textbf{Multi-view Ensembling}: Exploiting multi-view representation for 3D visual grounding has been studied before. As mentioned in the main manuscript, MVT~\cite{huang2022multiview}, ViewRefer~\cite{guo2023viewrefer} and Multi3DRefer~\cite{zhang2023multi3drefer} aggregate 3D, textual or 2D features from multiple views to reduce dependence on a specific viewpoint. Thus, each of the aforementioned works requires multiple forward passes of their respective encoder to extract multi-view features, resulting in a trade-off of performance versus run-time efficiency. Compared to the aforementioned works, our multi-view ensembling (MVE) directly operates on selected referred objects rather than each individual predicted object. This means that not only does MVE require multiple forward passes of the 3D encoder but also the decoder to extract mask information. While this is a notable limitation of the module, we would like to provide further advantages and counterpoints that emerge from such a limitation:}
\begin{enumerate}
    \item The parallelization of multi-forward pass methods for real-world applications is accomplished by constructing multiple copies of a single model. Essentially this maps the trade-off of performance versus run-time efficiency to that of memory efficiency. Given that the majority of the model's complexity stems from the encoder, the additional decoder forward passes required by MVE compared to existing approaches only incur minimal additional memory costs.
    \item Previous methods propose decoders specifically designed to handle multi-view features to tackle 3D visual grounding, while MVE only operates on the outputs of a single model. Thus MVE is a more flexible contribution as it can be utilized with any dense 3D visual grounding model, under the assumption that the model employs a grounding-by-selection approach.
\end{enumerate}

\end{document}